\PassOptionsToPackage{pdftex,dvipsnames}{xcolor}
\documentclass[letterpaper, 10 pt, conference]{ieeeconf}  

\IEEEoverridecommandlockouts                             
\usepackage{graphics}
\usepackage{epsfig} 
\usepackage{mathptmx} 
\usepackage{times} 
\usepackage{amsmath} 
\usepackage{amssymb}  
\usepackage{subcaption}
\usepackage{graphicx}
\usepackage{float}
\usepackage{bm}
\usepackage{caption}
\usepackage{mathtools}
\usepackage{stfloats}
\usepackage{mathrsfs}
\usepackage{multirow}
\usepackage{makecell}
\usepackage{vcell}
\usepackage{booktabs}
\usepackage{diagbox}
\usepackage{csquotes}
\usepackage{cite}
\usepackage{tikz}
\usepackage{lipsum}
\usepackage{schemata}
\usepackage{amsmath}
\usepackage{amssymb}

\usepackage[hyphens]{url}
\usepackage[hidelinks]{hyperref}
\hypersetup{breaklinks=true}
\usepackage{cite}

\usepackage{algorithm}
\usepackage[noend]{algpseudocode}

\algrenewcommand\algorithmicforall{\textbf{for each}}
\algrenewcommand\algorithmicindent{.8em}

\usepackage{hyperref}
\hypersetup{
colorlinks=true,
linkcolor=blue,
filecolor=magenta,
urlcolor=blue,
}

\usepackage[font=footnotesize]{caption}

\title{\LARGE \bf
ZeroCAP: Zero-Shot Multi-Robot Context Aware Pattern Formation via Large Language Models
}
\author{Vishnunandan L. N. Venkatesh, and Byung-Cheol Min
\thanks{This material is based upon work supported by the National Science Foundation under Grant No. IIS-1846221. Vishnunandan L. N. Venkatesh and Byung-Cheol Min are with the SMART Lab, Department of Computer and Information Technology, Purdue University, West Lafayette, IN 47907, USA
{\tt\small \{lvenkate,minb\}@purdue.edu}}%
}

\begin{document}
\maketitle
\thispagestyle{empty}
\pagestyle{empty}

\begin{abstract}
Incorporating language comprehension into robotic operations unlocks significant advancements in robotics, but also presents distinct challenges, particularly in executing spatially oriented tasks like pattern formation. This paper introduces ZeroCAP, a novel system that integrates large language models with multi-robot systems for zero-shot context aware pattern formation. Grounded in the principles of language-conditioned robotics, ZeroCAP leverages the interpretative power of language models to translate natural language instructions into actionable robotic configurations. This approach combines the synergy of vision-language models, cutting-edge segmentation techniques and shape descriptors, enabling the realization of complex, context-driven pattern formations in the realm of multi robot coordination. Through extensive experiments, we demonstrate the systems proficiency in executing complex context aware pattern formations across a spectrum of tasks, from surrounding and caging objects to infilling regions. This not only validates the system's capability to interpret and implement intricate context-driven tasks but also underscores its adaptability and effectiveness across varied environments and scenarios. The experimental videos and additional information about this work can be found at \href{https://sites.google.com/view/zerocap/home}{https://sites.google.com/view/zerocap/home}. 
\end{abstract}

\section{Introduction}
\label{sec:intro}

\begin{figure}[t]
\centering
\vspace{0pt}
    \includegraphics[width=0.95\linewidth]{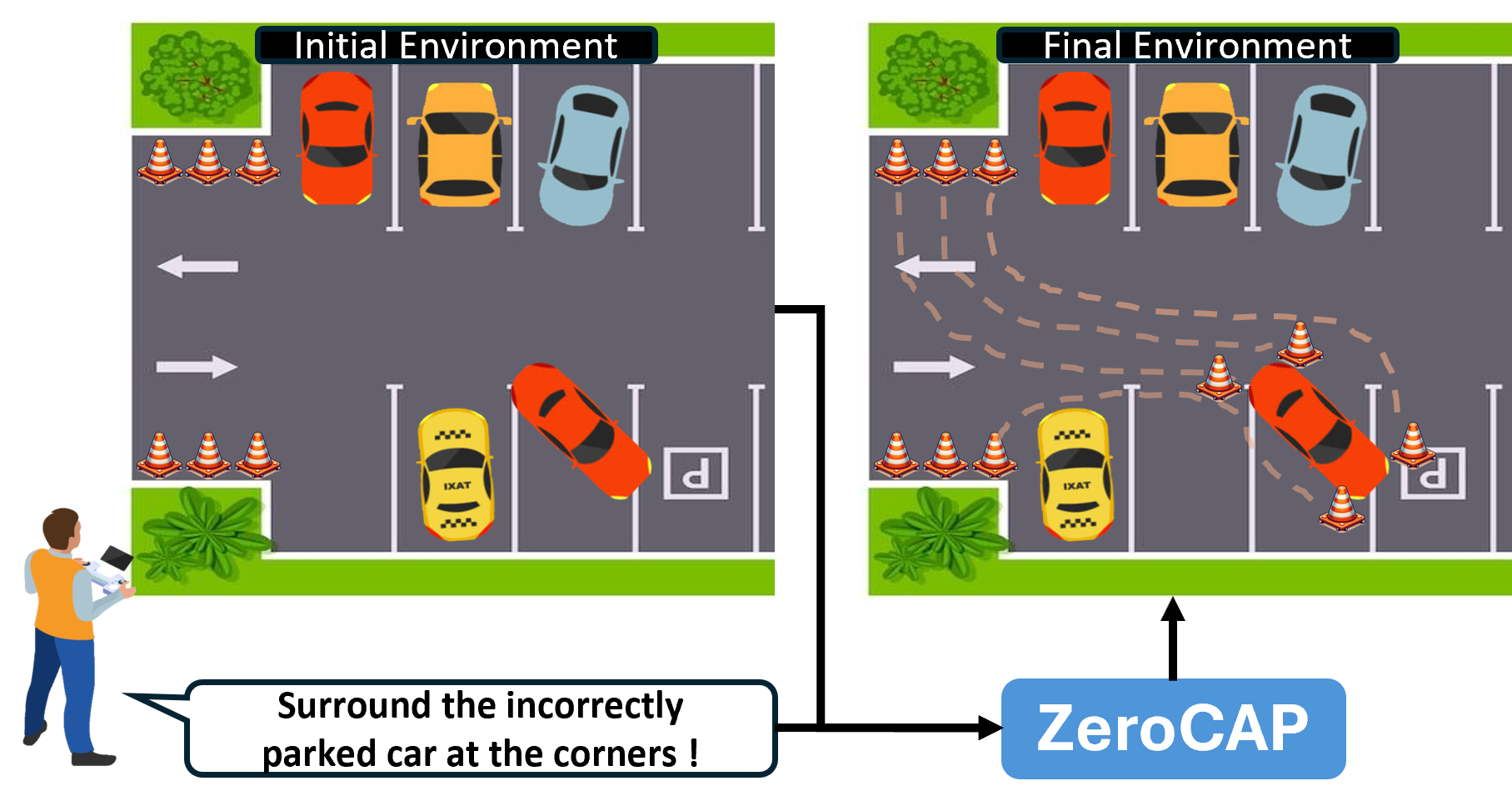}
    \vspace{-5pt}
\caption{A conceptual image illustrating the proposed \textbf{Zero}-Shot multi-robot \textbf{C}ontext \textbf{A}ware \textbf{P}attern (ZeroCAP) formation system, beginning with an initial environment where a car is parked incorrectly. Upon receiving the command from human operator to ``Surround the incorrectly parked car at the corners!", the system identifies the target vehicle and autonomously positions the robot cones at strategic corner locations, showcasing the successful execution of a zero-shot context aware pattern formation task.}
\label{fig:intro_pic}
   \vspace{-17pt}

\end{figure}

Pattern formation, a key idea in the organization of multi-robot systems (MRS) \cite{oh2015survey}, involves arranging robots in a specific way to reach a common objective. This ability is critical for applications such as environmental surveillance, precision farming, cooperative transportation, and emergency management. Its value lies in coordinating robots to perform tasks beyond the capabilities of individuals, enhancing effectiveness, adaptability, and robustness.  In this paper, we go beyond traditional geometric arrangements, focusing on purposeful pattern formations-where patterns are closely tied to user provided contexts and environments. This approach marks an evolution in MRS, steering the field towards more interactive and object-centric tasks that can be applicable to real world scenarios as shown in Fig. \ref{fig:intro_pic}. 

Despite its significance, traditional approaches to pattern formation in robotics have been predominantly constrained by model/rule-based methods and machine learning paradigms. These methodologies, while foundational, have their limitations:  model/rule-based strategies \cite{oh2015survey,xu2016clustered,5980269,gayle2009multi} are often scenario-specific and lack adaptability to new patterns, while reinforcement learning (RL) requires significant training and reward design, making it impractical for immediate deployment. Moreover, machine learning (ML) methods \cite{lin2023clip,8999149, prasad2017multi} rely on large, annotated datasets, which are challenging in dynamic environments. These limitations are especially evident when pattern formation must be conditioned on user-provided natural language inputs referring to specific objects, emphasizing the need for a more flexible and intuitive approach.

Integrating language comprehension into robotic systems offers considerable progress \cite{wu2023tidybot,10161317}, but it also presents challenges in spatial tasks like pattern formation. While language models have shown a remarkable ability as general pattern machines \cite{mirchandani2023large} and offer a robust framework for parsing and generating human-like text, Vision Language Models (VLM) face challenges in precise spatial localization tasks due to limitations in spatial reasoning and dataset training \cite{cai2024spatialbot,zhang2024vision}. This underscores the need for approaches that combine abstract instruction interpretation with accurate spatial execution by robots.

Our contribution to this evolving field addresses the challenge previously identified by \cite{5980269}, focusing on the dynamic assignment of a variable number of robotic agents to specific target positions for intricate pattern formations. By combining vision-based processing with the sophisticated interpretative capabilities of Large Language Models (LLMs), we introduce a \textbf{Zero}-Shot multi-robot \textbf{C}ontext \textbf{A}ware \textbf{P}attern (ZeroCAP) formation system that facilitates zero-shot, language, and context-conditioned pattern formation without any prior training or data. It processes both images and natural language instructions to identify objects of interest and generate precise coordinates for robot deployment, enabling immediate and purposeful pattern formation directly influenced by user instructions.

The main contributions of this paper are as follows: 
\begin{itemize} 
\item ZeroCAP, a novel zero-shot multi-robot pattern formation framework that integrates LLMs with vision-based processing to translate user natural language inputs into context-aware, purposeful pattern formations, moving beyond traditional geometric arrangements.

\item Implementation and evaluation of the framework in both simulation and real-world environments, validating the system's ability to perform complex pattern formations with varied objects and natural language inputs. \end{itemize}

The following sections will explore the challenges tackled, methodologies employed, and the broader implications of our work for advancing intuitive, flexible, and context-aware MRS.

\vspace{-5pt}
\section{Related Works}
\label{sec:rel_work}

\noindent\textbf{Pattern Formation.} Multi-robot pattern formation has been explored through both centralized \cite{belta2002trajectory}, \cite{indelman2018towards} and distributed methods \cite{oh2015survey}, \cite{cao2012overview}. Centralized systems, while efficient in task coordination, can suffer from single points of failure and may be less adaptable in highly dynamic environments. Distributed approaches, such as consensus-based \cite{olfati2007consensus}, leader-follower \cite{hong2008distributed}, and bio-inspired methods \cite{liu2019distributed}, improve adaptability and eliminate single points of failure but often rely on rule-based strategies that lack flexibility when applied to diverse or complex tasks. Advancements like potential field methods \cite{belta2004abstraction}, \cite{gayle2009multi}, and event-triggered mechanisms \cite{xu2016clustered} have been proposed, yet rule-based models still lack adaptability to novel or unseen patterns, while approaches like Mora et al. \cite{5980269}, emphasize visually appealing, artistic robot movements. Learning-based strategies like Q-learning \cite{prasad2017multi}, \cite{8999149}, while more generalizable, require extensive training and are limited to fixed patterns. Despite progress, challenges remain in developing flexible solutions that quickly adapt to user-specific contexts without pre-configuration or heavy training, moving beyond traditional geometric patterns to more contextually relevant formations.

Our approach bypasses these limitations by employing zero-shot learning with a language model, enabling flexible, intuitive multi-robot pattern formation. Leveraging natural language instructions, our system adapts to novel static scenarios without extensive training or predefined rules, marking a significant improvement over existing methods.

\noindent\textbf{LLM for Robotics.} LLMs are pivotal in robotics for task planning through zero-shot learning \cite{brown2020language,madaan2022language,mirchandani2023large}, excelling in diverse plan generation \cite{lin2023text2motion,10161317,wu2023tidybot} and performance refinement \cite{vemprala2023chatgpt}.  In the realm of multi-agent systems, LLMs have been instrumental in fostering agent cooperation, equipping individual agents with the capabilities to improve interaction dynamics and collaborative efficacy \cite{liu2023bolaa,hong2023metagpt,kannan2023smartllm}. Despite their growing utility in robotics, none of these noteable methodologies cover pattern formation among MRS. 

VLM, while useful in many applications involving robot navigation\cite{pmlr-v205-shah23b} and high-level action control\cite{brohan2023rt,gao2023physically}, are limited by their spatial reasoning capabilities \cite{zhang2024vision}, impacting their effectiveness in robotics applications that require precise spatial coordination. Our research leverages VLMs and LLMs for multi-robot pattern formation, addressing the spatial task limitations of VLMs and pioneering a natural language-driven approach to complex robotic coordination aided by representing shapes using a node graph.
\label{sec:rel_lfd}
  
\vspace{-5pt}
\section{Problem Description}
\label{sec:problem}

This research focuses on the task of deploying robotic agents to form patterns around or inside a designated object of interest ($O$) within an environment ($e$), depicted through an image ($I_E$) and guided by a natural language instruction ($Ins$). We operate under a centralized control system where robots are simplified as point masses and coordinated to achieve precise pattern formation based on $O$ and $Ins$. This centralized approach is adopted to streamline task coordination by
eliminating the need for each robot to independently process
language and spatial data. It reduces computational overhead
and avoids the significant costs of equipping each robot with
its own VLM/LLM.

The procedure begins with the identification and segmentation of $O$ from $I_E$, leading to a mask image that segregates $O$ from the rest of $e$. This step is crucial to accurately define the geometric shape of $O$. The shape of $O$ is mathematically modeled as a connected graph $G = (V, E)$ \cite{qureshi2007graph}, where $V$ signifies the set of vertices or nodes, outlining discrete points that mark the shape's corners or distinct features, and $E$ represents the set of edges connecting these vertices, thus framing the pattern without self-connectivity.

Let $R = \{r_1, r_2, \ldots, r_N\}$ represent the list of robotic agents available, where $N$ is the total number of robots. Given $Ins$, and $I_E$, the system, through a series of steps that include object detection, segmentation, and shape determination resulting in $G$, leverages an LLM to interpret and reason natural language instructions in light of $G$. The output of the system is a list of $(x, y)$ coordinates, $C_{xy} = \{(x_1, y_1), (x_2, y_2), \ldots, (x_N, y_N)\}$, which assigns each robot $r_i$ in $R$ to a specific coordinate $(x_i, y_i)$ for deployment. This assignment facilitates the formation of the specified pattern around/inside $O$, in accordance with $Ins$. While the current formulation operates in 2D, based on the 2D input image $I_E$, it can be generalized to n-dimensions. 

Formally, this process can be encapsulated as a function $F$, where $F(I_E, Ins) \rightarrow C_{xy}$, mapping the inputs to the deployment coordinates for each robot. This function represents the ZeroCAP system, which is detailed further in Sec. \ref{sec:methodology}.

 \begin{figure*}
 \vspace{0pt}
  \centering
  \includegraphics[width=0.98\textwidth]{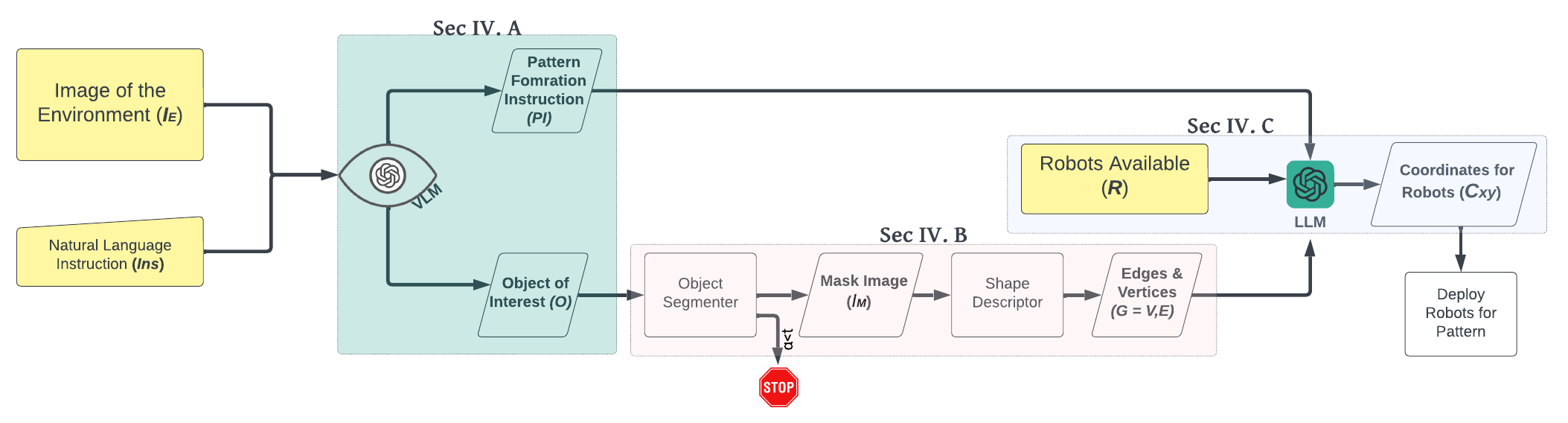} 
  \vspace{-10pt}
    \caption{An overview of the ZeroCAP system. It traces the workflow from the initial natural language instruction and input image of the environment, to the final deployment of robots, illustrating the sequence of processing stages—including context identification using  Vision Language Model
(VLM), object segmentation, shape description, and Large Language Model (LLM) coordination for precise robot placement in the environment. Three key stages are highlighted and explained in Section \ref{sec:methodology}.}
   \label{fig:overall_system}
   \vspace{-15pt}
\end{figure*}

\section{ZeroCAP Formation System}
\label{sec:methodology}

This section outlines the methodology behind the ZeroCAP formation system. The system facilitates zero-shot, context-aware pattern formation for MRS by integrating LLMs with visual perception and analysis. The system, represented by the function $F$, takes two key inputs: $I_E$ and $Ins$. The VLM processes these inputs to identify $O$ and generate a pattern formation instruction ($PI$). The system then segments the object, extracts shape descriptors, and the LLM uses this spatial descriptors along with the instruction to generate precise deployment coordinates for the robots. Finally, under centralized control, the robots execute the task based on the generated coordinates. The overall structure of the ZeroCAP framework is depicted in Fig. \ref{fig:overall_system}. Algorithm \ref{alg:alg} outlines the execution pipeline of the system, from processing the inputs to deploying robots to the designated coordinates. 

\subsection{Context Identification using VLM}
\label{sec:context}

This stage, represented as box `Sec IV. A' in Fig. \ref{fig:overall_system}, begins with a VLM interpreting the environmental image $I_E$ and $Ins$. The VLM is proficient at correlating visual and linguistic inputs, emulating human-like proficiency in image-language tasks. This stage produces two key outputs: $O$ and $PI$.

\begin{figure}
\centering
    \includegraphics[width=1\linewidth]{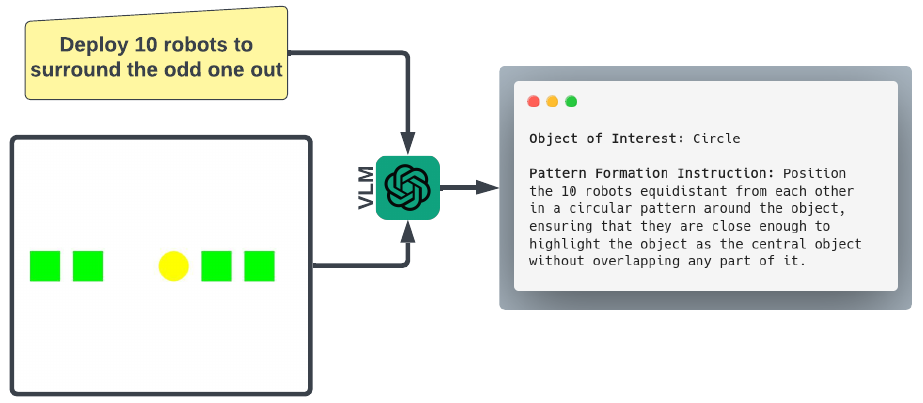}
    \vspace{-15pt}
\caption{An illustration of the VLM processing a natural language instruction to identify the object of interest and generate a pattern formation instruction, which guides the deployment of robots in a specified arrangement around the object. The object is not explicitly mentioned in the instruction and must be reasoned by the VLM within the context of image of the environment. }
\label{fig:context_pic}
   \vspace{5pt}
\end{figure}

$O$ is detected using the VLM's zero-shot capabilities, eliminating the need for prior object-specific training and allowing versatility in diverse environments. Unlike conventional object detection, which may localize objects but lacks contextual reasoning, the VLM can reason about the context and identify complex instructions \cite{zhang2024vision}. Taking the case of Fig.~\ref{fig:context_pic}, standard natural language-aided object detectors~\cite{lin2023clip} struggle to reason about which object is odd, whereas the VLM's advanced reasoning, guided by the user instruction, makes it more effective at identifying objects in complex scenarios. This advanced reasoning enables the VLM to align object identification with the given language directives. 

Simultaneously, the $PI$ is derived from the natural language directive, enriched by the context in $I_E$. Here, the VLM generates $PI$, inspired by chain-of-thought reasoning \cite{wei2023chainofthought}, to discern spatial and context-sensitive task directives. Algorithm \ref{alg:alg}, Line 1 corresponds to estimating the context using the VLM.  

One key strength of our system is addressing the challenge of precise spatial localization, a known limitation of standard VLMs. VLMs are not well-suited to directly generate $C_{xy}$, so instead, they focus on accurately identifying the $O$ and extracting a context-rich $PI$. These outputs are crucial for subsequent segmentation and pattern formation, handled by more specialized subsystems below. By focusing on general features and context from the environment and language input, the VLM ensures accurate identification of $O$ for further processing.

The VLM's output, consisting of $O$ and $PI$, thus propels the ZeroCAP system into the segmentation phase, where $O$ is visually distinguished within $I_E$, and $PI$ informs the strategic pattern formation instruction to be implemented.

\subsection{Object Segmentation and Shape Description}
\vspace{-5pt}
\label{sec:segment}
Following the identification of $O$ and acquiring the $PI$ from the VLM, the system moves to segmentation and shape description represented as box `Sec IV. B' in Fig. \ref{fig:overall_system}. These stages utilize language-conditioned segmentation \cite{lin2023clip,kirillov2023segany}, and standard computer vision methods for shape analysis.

\subsubsection{Object Segmentation}
The segmentation process employs LangSAM, an open-source language-conditioned model based on the segment-anything approach and GroundingDINO \cite{kirillov2023segany,liu2023grounding}. Given $O$ and $I_E$, the model generates a binary segmentation map, highlighting the object in white against a black background. The segmentation outputs a Masked Image $I_M$ and an accuracy metric $\alpha$, which is a scalar value compared against a threshold $t$, a predefined scalar constant. If $\alpha$ exceeds $t$, the process continues, ensuring the segmentation quality is sufficient. This approach precisely extracts $O$ from $I_E$, encapsulated in $I_M$ for further processing (Algorithm \ref{alg:alg}, Lines 2-4).

\subsubsection{Shape Description} With a quality $I_M$, the system proceeds to shape description. Using standard computer vision techniques, $I_M$ is preprocessed for edge and contour detection through OpenCV, resulting in the identification of edges $E$ and vertices $V$. These form a connected graph $G = (V, E)$, which mathematically represents $O$. The process is shown in Algorithm \ref{alg:alg}, Lines 9-12, initiated by Line 5. 

ZeroCAP thus integrates segmentation and shape description to ensure accurate object recognition and modeling, forming the basis for the LLM-based Pattern Former.

\begin{algorithm}[t]
\vspace{10pt}
\caption{ZeroCAP System Pipeline}
\begin{algorithmic}[1]
\Require Image of the Environment $I_E$, Natural Language Instruction $\textit{Ins}$
\Ensure List of coordinates for robot deployment $C_{xy}$

\State $(O, PI) \gets \Call{VLM.Process}{I_E, \textit{Ins}}$
\State $(I_M, \alpha) \gets \Call{Segment}{O, I_E}$

\If{$\alpha < t$}
    \State \Return \textit{``Insufficient segmentation accuracy.''}
\EndIf

\State $(V, E) \gets \Call{ShapeDescriptor}{I_M}$
\State $C_{xy} \gets \Call{LLM.GenerateCoordinates}{PI, V, E, R}$

\For{each robot $r_i$ in $R$ and coordinate $(x_i, y_i)$ in $C_{xy}$}
    \State \Call{DeployRobots}{$r_i, (x_i, y_i)$}
\EndFor

\Procedure{ShapeDescriptor}{$I_M$}
    \State \textit{Preprocess the masked image $I_M$}
    \State \textit{Extract edges $E$ and vertices $V$}
    \State \Return $(V, E)$ \textit{from $I_M$}
\EndProcedure

\Procedure{DeployRobots}{$r, (x, y)$}
    \State \textit{Command robot $r$ to move to coordinates $(x, y)$}
    \State \textit{Verify that robot $r$ has reached $(x, y)$}
\EndProcedure
\end{algorithmic}
\label{alg:alg}
\end{algorithm}
\setlength{\textfloatsep}{0.15cm}

\vspace{-5pt}
\subsection{LLM-based Pattern Former}
\vspace{-4pt}
\label{sec:pattern}
After delineating the geometry of $O$, the system advances to generating goal position coordinates using a LLM represented as box `Sec IV. C' in Fig. \ref{fig:overall_system}. The LLM receives the number of available robots $R$, $PI$, and the spatial characteristics of the object, represented by its edges \(E\) and vertices \(V\), to compute the optimal robot deployment positions. Instead of using the entire Masked Image \(I_m\), we opt for the more computationally efficient use of \(E\) and \(V\), significantly reducing token overhead for LLMs and enhancing real-time performance. This approach ensures faster processing while keeping computational costs low.

The LLM integrates the $PI$ with the spatial data \(E\) and \(V\) to determine the optimal pattern formation. This involves comprehensive reasoning to align the robots according to $PI$, while considering the shape of the object and the available robotic agents $R$. The output of this stage is a list of coordinates \(C_{xy}\) specifying precise deployment locations for each robot, as outlined in Line 6 of Algorithm \ref{alg:alg}.


\vspace{-10pt}
\subsection{Robot Deployment}
\vspace{-2pt}
Once the deployment coordinates \(C_{xy}\) are generated, the ZeroCAP system initiates robot deployment, translating these coordinates into coordinated actions.

Each robot \(r_i, i \in N\) is directed toward its assigned location \((x_i, y_i)\). Upon arrival, the system verifies each robot's position against \(C_{xy}\), making adjustments as needed to correct any deviations, ensuring the formation aligns with the pattern instructions. This is shown in Lines 7 to 8 and 13 to 15 in Algorithm \ref{alg:alg}.

The successful deployment of robots around the $O$ completes the pattern formation task, demonstrating ZeroCAP’s ability to execute context-aware, instruction-based patterns, without the need for any prior training.

\vspace{-2pt}
\section{Experiment}
\vspace{-2pt}

 \begin{figure*}
  \centering
  \includegraphics[width=\textwidth]{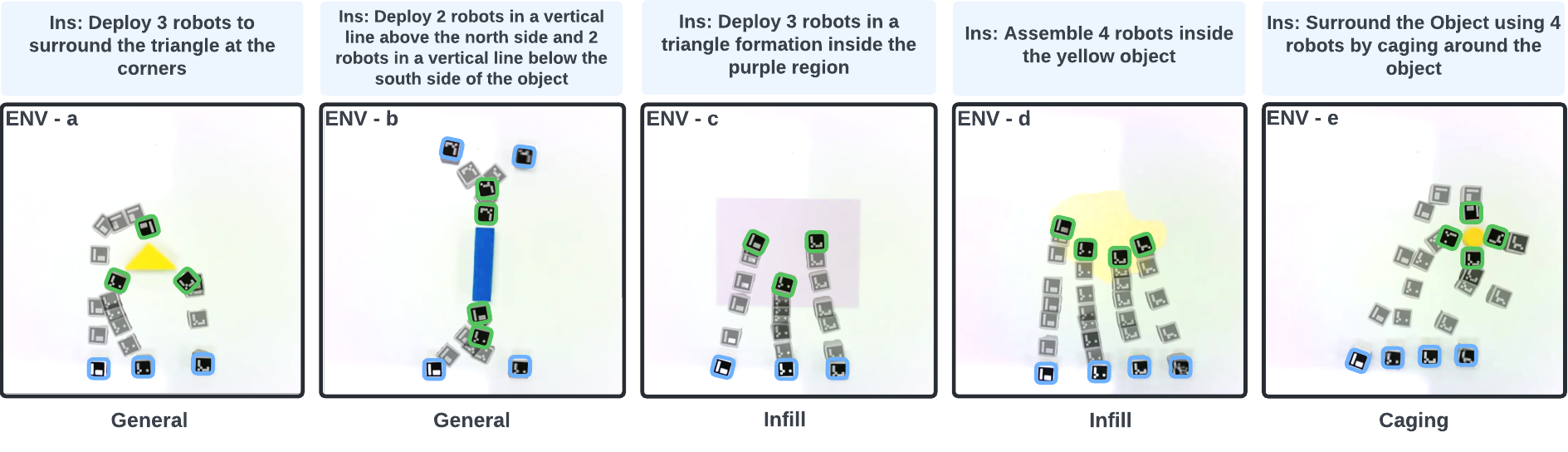} 
  \vspace{-20pt}
    \caption{Illustrations of five real-world tasks performed by the ZeroCAP system, showcasing different pattern formations: (a) and (b) illustrate general pattern formations; (c) and (d) depict infill pattern formations; (e) demonstrates a caging task. Each task is executed based on natural language instructions provided at the top of each subfigure. Blue boundaries around the robots indicate their initial positions, while green boundaries represent their final positions. Please zoom in on the images for details.}
   \label{fig:pattern_formation_tasks}
   \vspace{-15pt}
\end{figure*}

We evaluate the ZeroCAP system's zero-shot pattern formation capabilities across five real-world tasks, as shown in Fig. \ref{fig:pattern_formation_tasks} and ten tasks performed in simulated environments. These tasks are categorized into three types: 1) \textit{General pattern formation}, where robots arrange around the object of interest; 2) \textit{Infill pattern formation}, where robots occupy the interior of a region; and 3) \textit{Caging formation}, where robots create a boundary around the object. This categorization shifts from standard geometric patterns to contextually relevant formations, driven by natural language instructions and environmental cues, demonstrating the system’s flexibility.

\begin{figure}[t]
\centering
\vspace{-5pt}
    \includegraphics[width=0.97\linewidth]{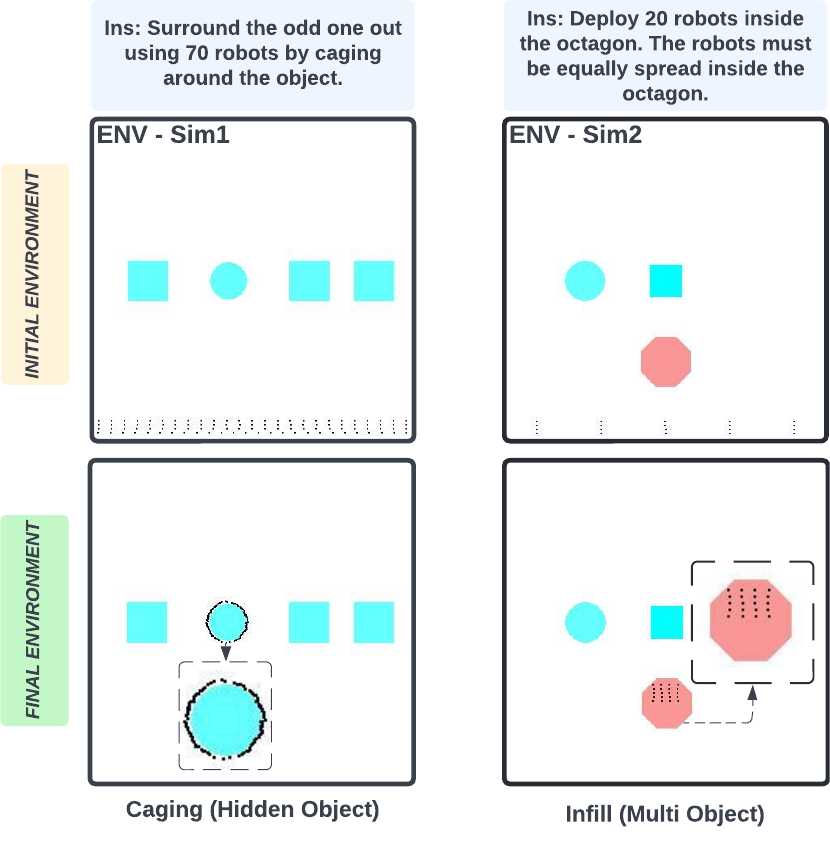}
    \vspace{-10pt}
\caption{Two simulation scenarios demonstrating ZeroCAP's ability to execute context-aware pattern formations. Sim1 (left) depicts a caging task in a hidden object setup, where the object of interest is not directly specified and must be inferred. Sim2 (right) illustrates an infill task in a multi-object setup. Please zoom in on the images for details.}
\vspace{-5pt}
\label{fig:sim_pic}
\end{figure}

\begin{figure}[]
\centering
    \includegraphics[width=\linewidth]{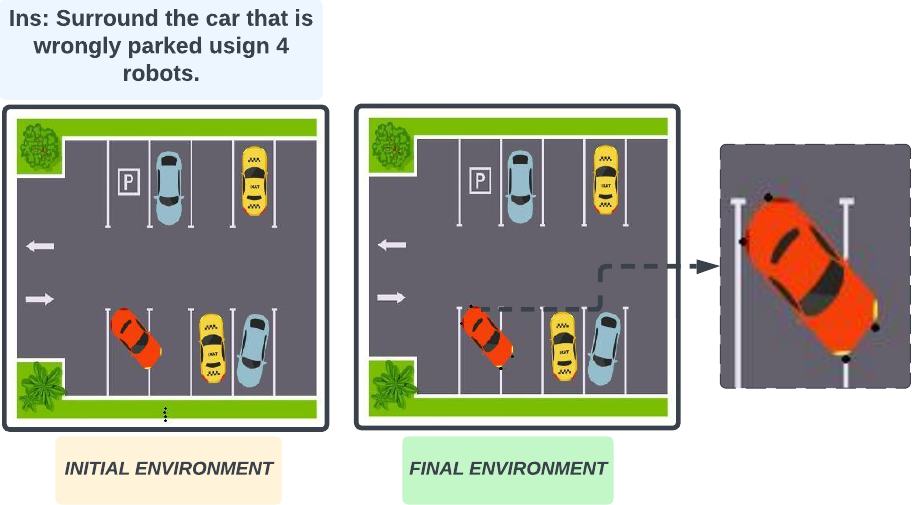}
    \vspace{-10pt}
\caption{Illustration of a general pattern formation task as shown in concept Fig. \ref{fig:intro_pic}, featuring a hidden object setup where the system must infer the wrongly parked car and deploy robots, represented by black points, to surround it.}
\label{fig:final_park}
\end{figure}

The tasks also vary by object setup to test adaptability. In the \textit{Single Object} setup as shown in Fig. \ref{fig:pattern_formation_tasks}, a single object simplifies identification. The \textit{Multi-Object} setup, shown in Fig. \ref{fig:sim_pic} adds distractors, requiring the system to isolate the target from several options based on explicit instructions. In the \textit{Hidden Object} setup, shown in Fig. \ref{fig:sim_pic} and Fig. \ref{fig:final_park}, the target is implied rather than stated, requiring the system to reason based on contextual clues.

\vspace{-3pt}
\subsection{Baseline Comparisons}

The system is benchmarked against three baselines: an RL-based approach with hand-crafted rewards for pattern formation—where Caging rewarded enclosing a target, Infill incentivized structured spacing inside a region, and General used a proximity-based reward for clustering robots near a reference point—and two vision-language models, GPT-4 Vision \cite{openai2023gpt4v} and LLava-v1.6 \cite{liu2023llava}. Experiments employ GPT-4 \cite{openai2023gpt4}, LLaMA-70B \cite{touvron2023llama}, and Claude \cite{anthropic}, with GPT-4 serving as the default in ablation studies. Unlike traditional rule-based pattern formation methods, ZeroCAP interprets natural language instructions to enable context-aware formations, rendering direct comparisons with classical approaches infeasible due to their lack of user intent integration.

\subsection{Environment}
Experiments were conducted in both real-world settings and in 2D simulations. Real-world experiments used a testbed with Hamster mobile robots under an overhead camera \cite{lee2021investigating}, with the environment simplified into a grid-based image to ensure consistency in point-mass robot deployment. Fig. \ref{fig:pattern_formation_tasks} highlights the real-world tasks, demonstrating the practical viability of our approach. Additionally, 2D simulations were carried out using the Pygame environment, as shown in Fig. \ref{fig:sim_pic}, where robots were modeled as point-mass entities and moved precisely to target coordinates. The simulation setup included multi-object tasks with infill and caging requirements, excluding obstacle avoidance to focus solely on pattern formation. These simulation experiments served to rigorously validate the proposed methods across varied conditions and object setups. For more detailed demonstrations of the system under various setups and task categories, please refer to the supplementary video.

\subsection{Evaluation Metrics} 
We used two evaluation metrics: Success Rate (SR) and Goal Condition Recall (GCR). SR indicates if the robots successfully formed the pattern by comparing final positions to handcrafted ground truth labels, with success requiring all positions to match. GCR measures the proportion of correctly positioned robots, with a GCR of 1 implying an SR of 1. Each task was run over 10 trials, and average SR and GCR were reported to account for LLM variability.

\vspace{-2pt}
\section{Results and Analysis}

Table \ref{tab:results} summarizes the evaluation of the tested methodologies, highlighting ZeroCAP's effectiveness over the baselines. GPT-4 Vision and LLava v1.6, are tested for direct VLM-only zero-shot pattern formation, both performed poorly. GPT-4 Vision failed to generate meaningful coordinates, revealing its limitations in spatial reasoning. LLava produced inaccurate or irrelevant positions. These results show that using VLMs alone for context-aware pattern formation is inadequate without enhanced spatial reasoning capabilities. The RL-based baseline showed mixed results, struggling with general and infill pattern formation due to varied user instructions requiring flexibility. RL's reliance on predefined rewards limits generalizability; while effective in task-specific training, it underperforms with generalized reward structures. We chose a generalized reward approach to enhance scalability, as designing specific rewards for every new task is impractical. Consequently, RL performed poorly in general and infill tasks but excelled in caging tasks, where the consistent goal of forming a barrier aligned with the reward design. This highlights RL’s strength in structured tasks but its limitations in adapting to nuanced, context-driven user instructions.

In contrast, the ZeroCAP system, leveraging LLMs like GPT-4, llama-2-70b, and Claude-3-opus, significantly outperformed the baselines. GPT-4-based ZeroCAP achieved the highest SR and GCR across all tasks, largely due to its superior language processing, allowing for better interpretation of complex instructions. The ZeroCAP system’s architecture, which decouples spatial reasoning from the VLM and utilizes edge representations, mitigated GPT-4 Vision's limitations in direct spatial localization. Llama-2-70b performed moderately well but showed inconsistencies, particularly in tasks with hidden object scenarios requiring complex spatial reasoning. Claude-3-opus, while slightly less accurate than GPT-4, demonstrated more consistency across different task types.

\vspace{0pt}

Task-wise, general pattern formation was the most challenging due to language ambiguity and the lack of structured references, leading to misinterpretations in robot placements. Infill tasks achieved the highest SR and GCR, though occasional failures occurred from uneven spacing inside the region. Caging tasks, while more structured, encountered difficulties with precise localization, often resulting in incomplete barriers due to gaps in robot placement, but ZeroCAP managed consistent barrier formations effectively.

\begin{table*}[t]
\centering
\caption{Comparative analysis of pattern formation methods for General, Infill, and Caging tasks, evaluating Success Rates (SR) and Goal Condition Recall (GCR) metrics across different methods. Total represents the combined performance across all task categories. Bold numbers indicate the best performance among the tested methods.} 
\vspace{-5pt}
\label{tab:results}
\resizebox{0.8\textwidth}{!}{
\tiny
\begin{tabular}{lcccccccc}
\toprule
\multirow{2}{*}{\textbf{Methods}} & \multicolumn{2}{c}{\textbf{General}} & \multicolumn{2}{c}{\textbf{Infill}} & \multicolumn{2}{c}{\textbf{Caging}} & \multicolumn{2}{c}{\textbf{Total}} \\
\cmidrule(lr){2-3} \cmidrule(lr){4-5} \cmidrule(lr){6-7} \cmidrule(lr){8-9}
                & \textbf{SR} & \textbf{GCR} & \textbf{SR} & \textbf{GCR} & \textbf{SR} & \textbf{GCR} & \textbf{SR} & \textbf{GCR} \\
\midrule
\textit{GPT-4 Vision (Only)}     & 0    & 0    & 0    & 0    & 0    & 0    & 0    & 0    \\
\textit{LLaVA v1.6 (Only)}       & 0    & 0    & 0 & 0.15  & 0    & 0.07 & 0    & 0.22 \\
\textit{RL Baseline}             & 0.35 & 0.53 & 0.28 & 0.41 & \textbf{0.88} & \textbf{0.9}  & 0.5  & 0.61 \\
\textit{ZeroCAP (GPT-4)}         & \textbf{0.71} & \textbf{0.85} & \textbf{0.86} & \textbf{0.89} & 0.74 & 0.82 & \textbf{0.77} & \textbf{0.85} \\
\textit{ZeroCAP (llama-2-70b)}    & 0.39  & 0.47 & 0.55 & 0.71 & 0.62 & 0.74 & 0.52  & 0.64 \\
\textit{ZeroCAP (claude-3-opus)} & 0.63 & 0.78 & 0.79 & 0.88 & 0.69 & 0.79 & 0.70 & 0.81 \\
\bottomrule
\end{tabular}}
   \vspace{-10pt}
\end{table*}

\vspace{-3pt}
\subsection{Ablation: Segmentation Methods}
Table \ref{tab:abs} presents the results of an ablation study comparing the performance of CLIPSeg \cite{lin2023clip} and LangSAM as standalone segmentation methods and their integration within the ZeroCAP framework. While CLIPSeg and LangSAM achieve modest success rates (0.51 and 0.43, respectively) when used alone, their performance is limited by a lack of spatial reasoning and deeper contextual understanding, particularly in tasks requiring hidden object identification.

\vspace{0pt}
\begin{table}[h]
\centering

\caption{Ablation studies comparing the success rates (SR) of standalone segmentation methods and their integration within the ZeroCAP system.}
\vspace{-0pt}
\label{tab:abs}
\begin{tabular}{lc}
\hline
\multicolumn{1}{c}{\textbf{Method}} & \textbf{SR} \\ \hline
\textit{CLIPSeg (Only)}             & 0.51        \\
\textit{LangSAM (Only)}             & 0.43        \\
\textit{ZeroCAP with CLIPSeg}       & 0.78        \\
\textit{ZeroCAP with LangSAM}       & 0.77        \\ \hline
\end{tabular}
\vspace{0pt}
\end{table}

When integrated into ZeroCAP, success rates rise to 0.78 and 0.77, respectively. This improvement demonstrates the added value of incorporating VLMs into the system, providing the necessary contextual reasoning and spatial capabilities to enhance overall performance.

\vspace{-5pt}
\subsection{Ablation: Shape Descriptors}

The ablation study presented in Table \ref{tab: shapeabl} examines the impact of different shape descriptors on ZeroCAP's performance. `Increase in Tokens' reflects the percentage increase in token usage compared to using only `Edges' as the shape descriptor.
\vspace{-5pt}
\begin{table}[h]
\centering

\caption{Ablation studies on the impact of different shape descriptors on Success Rates (SR) and their relative increase in tokens required for LLM processing.}
\vspace{-5pt}
\label{tab: shapeabl}
\begin{tabular}{lcc}
\hline
\multicolumn{1}{c}{\textbf{Shape Descriptor}} & \textbf{SR} & \multicolumn{1}{l}{\textbf{\begin{tabular}[c]{@{}l@{}}Increase \\ in Tokens\end{tabular}}} \\ \hline
\textit{Edges}                                & 0.77        & -                                                                                          \\
\textit{Edges and Vertices}                   & 0.48        & 2\%                                                                                       \\
\textit{Binary Image Matrix}                  & 0.71        & 400\%                                                                                      \\ \hline
\end{tabular}
   \vspace{-5pt}

\end{table}

Results show that using `Edges' alone yields the highest SR (0.77), demonstrating that this descriptor provides sufficient information for the LLM without increasing token load. Adding `Vertices' results in only a 2\% token increase, but the SR drops to 0.48, indicating that the complexity introduced by vertices can hinder performance, especially for non-polygonal shapes where vertex definitions are less clear. The binary image matrix, created by converting and resizing the object mask to a $100\times100$ grid, increases token usage by 400\%, making it less practical. While it achieves a decent SR of 0.71, the token cost and potential loss of important shape features during resizing highlight the trade-off between detail and efficiency in ZeroCAP's processing.

Overall, the results confirm that VLMs alone are insufficient for zero-shot context-aware pattern formation, underscoring the need for ZeroCAP. ZeroCAP outperforms baselines, particularly in tasks requiring nuanced reasoning. Ablation studies show that edges as shape descriptors yield the best success with minimal token load, highlighting ZeroCAP’s strength in integrating LLMs for robust, zero-shot pattern formation and setting a foundation for future multi-robot coordination research.

\section{Conclusion}
\label{sec:conclusion}

This work presents ZeroCAP, a framework integrating LLMs with MRS for zero-shot, context-aware pattern formation. By translating natural language instructions into precise spatial arrangements, ZeroCAP overcomes VLM spatial localization limitations by decoupling spatial reasoning from visual processing, using edge and vertex representations for more accurate pattern formation. Currently, as VLMs are mostly trained on 2D images \cite{cai2024spatialbot}, ZeroCAP is confined to a 2D framework for robustness, though future developments may extend it to 3D. The system is deployable with a map or overhead camera and could be expanded using multiple cameras in the future for wider coverage. While our focus has been on static formations, dynamic, real-time adaptations in complex, occluded environments hold significant promise, paving the way for more intuitive robotic coordination in applications such as surveillance and logistics.

\section*{Acknowledgement}
We would like to thank Dr. Tamzidul Mina at Sandia National Laboratories for his valuable feedback, which significantly enhanced the clarity and quality of this paper. 

\bibliographystyle{IEEEtran}
\newpage
\bibliography{references}
\end{document}